\documentclass{article}
\usepackage{graphicx} % Required for inserting images
\usepackage{parskip}
\usepackage[margin=4cm]{geometry}
\usepackage{caption}
\usepackage{float}
\usepackage{multirow}
\usepackage{color}
\usepackage{comment}
\usepackage{enumitem}
\usepackage{authblk}
\usepackage[hyphens]{url}
\usepackage{hyperref}
\usepackage[hyphenbreaks]{breakurl}

\title{A Survey of AI-Powered Mini-Grid Solutions for a Sustainable Future in Rural Communities}
\author[1]{Craig Pirie}
\author[1]{Harsha Kalutarage}
\author[1]{Muhammad Shadi Hajar}
\author[1]{Nirmalie Wiratunga}
\author[2]{Subodha Charles}
\author[2]{Geeth Sandaru Madhushan}
\author[2]{Priyantha Buddhika}
\author[2]{Supun Wijesiriwardana}
\author[2]{Akila Dimantha}
\author[2]{Kithdara Hansamal}
\author[2]{Shalitha Pathiranage}

\affil[1]{Robert Gordon University}
\affil[2]{AltaVision Solar}
\date{July 2024}

\begin{document}

\maketitle

\begin{abstract}
This paper presents a comprehensive survey of AI-driven mini-grid solutions aimed at enhancing sustainable energy access. It emphasises the potential of mini-grids, which can operate independently or in conjunction with national power grids, to provide reliable and affordable electricity to remote communities. Given the inherent unpredictability of renewable energy sources such as solar and wind, the necessity for accurate energy forecasting and management is discussed, highlighting the role of advanced AI techniques in forecasting energy supply and demand, optimising grid operations, and ensuring sustainable energy distribution.  This paper reviews various forecasting models, including statistical methods, machine learning algorithms, and hybrid approaches, evaluating their effectiveness for both short-term and long-term predictions. Additionally, it explores public datasets and tools such as Prophet, NeuralProphet, and N-BEATS for model implementation and validation. The survey concludes with recommendations for future research, addressing challenges in model adaptation and optimisation for real-world applications.
\end{abstract}

\section{Introduction}
Access to reliable and affordable electricity is a critical need for communities around the world, particularly in rural areas where access to the national power grid may be limited or non-existent. One potential solution to this problem is the use of mini-grids, which are small-scale electrical systems that generate and distribute electricity to a specific, often localised, geographical area. Mini-grids can be powered by various energy sources, including solar, wind, hydro, and diesel generators, and can be designed to operate in conjunction with or independent of the national power grid~\cite{IEA2018}. 

Mini-grids are a rapidly evolving technology due to the number of advantages they provide over typical national grid systems. Firstly, mini-grids can provide affordable and more reliable electricity to remote communities that have no connection, or an unreliable connection to the national grid. The rural nature of many communities means that mini-grid systems will often heavily rely on clean energy sources such as solar and wind, reducing reliance on depleting fossil fuels and thus improving environmental sustainability and health equity. In cases where communities do have access to the national grid, a local mini-grid will provide additional security and energy independence, and potentially provide the ability to sell excess energy to the national grid or neighbouring local grids~\cite{forecasting2018}. Overall, access to cheap, reliable, and cost-effective energy is a tremendous boon to rural communities, helping to stimulate economic development through job creation and increasing household income~\cite{iea2019world}. 

While prioritising the use of renewable energy sources is a noble and necessary goal, many clean energy sources suffer from a high degree of unpredictability due to how nature is inherently chaotic. Wind and solar energy sources, such as wind turbines and solar panels, are significantly affected by both short-term weather conditions and longer-term seasonal variations~\cite{forecasting2018}. Furthermore, the reliance on a mini-grid necessitates an in-depth understanding of the energy demand within the community, encompassing its fluctuations due to seasonal effects, human behaviour, and unforeseen adverse conditions~\cite{ma2013generic}. To ensure the smooth operation of a mini-grid, accurate forecasting of both energy generation and energy demand becomes crucial.

\section{Forecasting Techniques: Technical Analysis}

Forecasting is the process of predicting future events or trends based on historical data and other pertinent information. It is a critical tool for businesses, governments, and other organisations that need to plan for the future and make informed decisions. Forecasting is used in a wide range of applications, including finance, economics, weather forecasting, and supply chain management~\cite{TaylorLetham2018}.

Selecting an appropriate model is essential to achieving accurate forecasts. This can be a difficult decision due to the large number of models to choose from, each with its own strengths and limitations. The choice of model can depend on factors such as the nature of the data, the forecasting horizon, and the complexity of the relationships within the data. 

Typically, AI-driven mini-grid systems consist of three core components, each requiring accurate and timely forecasting:
\begin{itemize}
\item A forecast for the energy supply from all distributed energy resources connected to the mini-grid. Since the mini-grid will aim to be heavily reliant on renewable resources, a successful forecast will require accurate information on a variety of physical attributes, such as local weather patterns, temperature, cloud cover, and other unspecific atmospheric effects.
\item A forecast for the energy demand from the consumers of the mini-grid. This will require detailed knowledge of typical consumer behaviours and routines within the region, as well as information on upcoming important regional events.
\item A forecast for the optimal operation and management of the mini-grid is essential. Once the energy supply and demand are forecast, a model must determine the optimal approach to generating the required energy while meeting established objectives, such as minimising carbon emissions. The model must also forecast optimal times for grid management tasks, such as maintenance, based on energy forecasts.
\end{itemize}

The following subsections will detail existing research on the types of models and intelligent algorithms that could be utilised to achieve the aforementioned core components, emphasising their suitability.

\subsection{Statistical and Intelligent Techniques for Forecasting}

This review focuses on mini-grid systems that utilise renewable energy, with wind and solar as the primary distributed energy resources. Due to the inherently chaotic nature of weather and the strong dependence of renewable energy systems on current atmospheric conditions, forecasts for energy generation from renewable power sources usually fall within a short or \textit{very} short forecast horizon~\cite{forecasting2018}.

Techniques used within forecasting typically fall into one of four categories~\cite{soa, forecasting2018}: physical prediction models~\cite{hygro, energysavings, physical, physical2}, statistical models~\cite{chinama, nonlinear, italylr, armaxpv}, intelligent computational algorithms~\cite{svmconsumption, svmturkey,gaturkey, spainconsumption, cyprusconsumption}, and hybrid models~\cite{phsyical-stats, hybridhvac, gahvac}. 

The simplest type of forecasting model, typically used as a baseline for other models, is called the \textit{persistence} model. This model typically assumes that the value at the next time-step will simply be equal to the value at the current time-step~\cite{naive, dnaive}. This model is considered naive since it doesn't involve any intelligent or statistical techniques and makes a base assumption of continuity within the data. Slightly more complex persistence models can factor in the gradient of change of the time-series which produced the current time-step, and again assumes that the gradient will remain constant over time.

A physical forecasting model attempts to use the known laws of physics in order to simulate future atmospheric conditions, and the corresponding weather patterns, and thus forecast potential energy generation. The Numerical Weather Prediction (NWP) model is widely regarded as the prevailing technique for physical modelling within weather forecasting \cite{nwp, nwp2, nwp3}. The NWP model typically functions by dividing the atmosphere into a three-dimensional grid of cells, with each cell containing information about the temperature, pressure, wind speed, and humidity at that location. The model then relies on the physical theories of thermodynamics and fluid dynamics to simulate the way these variables will interact and change over time and predicts how the weather will evolve~\cite{ChallengesandOpportunitiesinNumericalWeatherPrediction}. Fundamental challenges to NWP models include the expertise in mathematics and physics required for their development and effective utilisation, alongside the required substantial computational power and resources, potentially restricting access for smaller or less developed countries~\cite{warner_2010, ChallengesandOpportunitiesinNumericalWeatherPrediction}. 

Statistical models are widely used techniques for forecasting time-series data due to their relative simplicity and ease of implementation~\cite{ma2018review}. These models typically utilise statistical techniques such as Auto-regression (AR) and Moving-averages (MA), or a combination of both (ARMA)~\cite{arma}. Auto-regression is a common time-series analysis method that aims to model the relationship between a time-series and its past values, utilising techniques such as Maximum Likelihood Estimation (MLE) to minimise the error between the predicted and actual values~\cite{forecasting2, forecasting2018}. Moving-averages involve taking an average of residuals over a moving time window in order to estimate the average size and direction of the errors in the forecast. By incorporating an integrated term into the ARMA model, the new ARIMA model~\cite{boxjenkins} can account for both the long-term trends or patterns in the data (represented by the AR term) and the random fluctuations around those trends (represented by the MA term), while also accounting for non-stationarity in the data. The result is a powerful modelling framework that can capture the complexity of time series data and produce accurate forecasts~\cite{fpp3}.

The use of deep learning techniques in forecasting models has gained popularity due to their capability to effectively handle complex and non-linear relationships between variables. These intelligent forecasting models also have the advantage of relying less on knowledge of any underlying physical processes, instead utilising a more \textit{black box} approach~\cite{forecasting2018}.  One of the most widely used AI techniques within forecasting includes the use of fuzzy logic~\cite{forecasting2018}.

Fuzzy logic can model complex relationships between input variables and forecast values. Within a fuzzy logic model, values are represented as a degree of membership to a linguistic category, rather than a simple \textit{crisp} value~\cite{ALI2018223, NeuralFuzzyInferenceSystem2014}. Fuzzy logic is useful for modelling relationships which are uncertain, imprecise, or difficult to quantify using traditional statistical methods, especially if the data is noisy or incomplete~\cite{RANAWEERA1996215}. Fuzzy logic works by first defining the input and output variables in terms of their fuzzy sets and membership functions. For example an input variable of \textit{temperature} could be defined by categories \textit{temperature\_low}, \textit{temperature\_medium}, and \textit{temperature\_high}, each of which is defined by a membership function. An example of such a variable with its member functions is shown in Figure \ref{fig:fuzzy1} below.

\begin{figure}[H]
\centering
\includegraphics[width=0.9\textwidth]{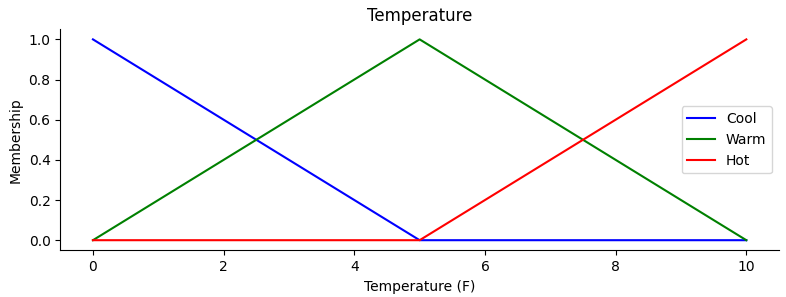}
\captionsetup{font=small}
\caption{An example of how the temperature variable can be defined by a fuzzy set of category memberships}
\label{fig:fuzzy1}
\end{figure}

Rules can then be established which define relationships between the input and output memberships. These rules typically take the form of \textit{if/then} statements such as \textit{IF temperature is high AND humidity is high THEN rain is high}. The rules are often generated using expert knowledge and can be combined to form a fuzzy inference system that can generate forecasts. Given new test data, the fuzzy inference system could provide a rain forecast by observing the degree of membership of the input values to each output category, or a process called defuzzification could be used to produce a crisp real-world value which could be better interpreted~\cite{RANAWEERA1996215}. Issues with simple fuzzy models stem from the rules and fuzzy sets which govern them. For complex systems, these rules and sets will require expert knowledge to craft, and an abundance of rules could easily lead to the model overfitting on any training data.

\begin{figure}[H]
\centering
\includegraphics[width=0.9\textwidth]{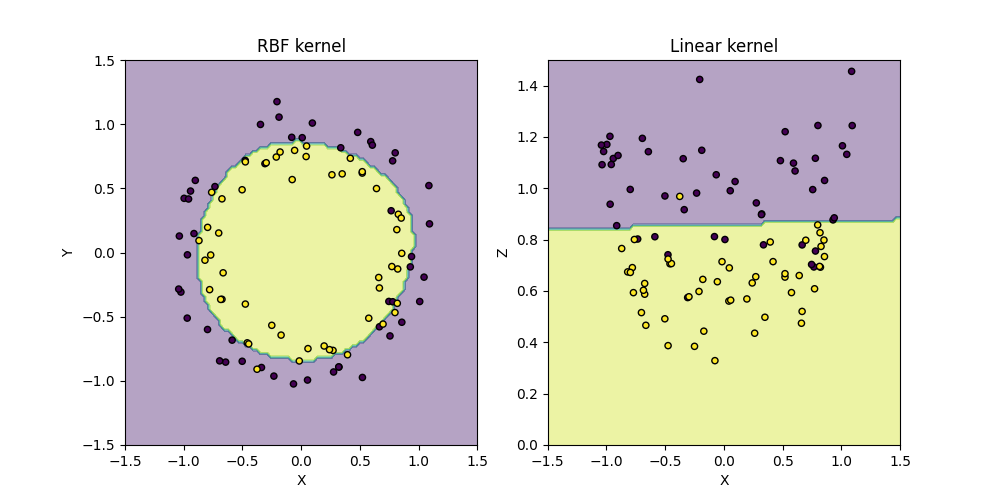}
\captionsetup{font=small}
\caption{Left: RBF kernel with non-linear boundary. Right: Linear kernel with linear boundary.}
\label{fig:SVM}
\end{figure}

Support vector machines (SVMs) are a type of supervised machine learning algorithm used for classification and regression analysis~\cite{Shmilovici2010}. SVMs work by attempting to produce a hyperplane, known as a decision boundary in classification problems, which separates the data into two or more classes based on their features. A hyperplane is simply a surface that is defined within the feature space which has the largest margin possible between the classes defined within the training data. To assist with finding the hyperplane a technique called the \textit{Kernel Trick} is often used. This trick aids in defining complex non-linear decision boundaries through the use of a kernel function which maps the input data onto a higher-dimensional feature space in which the data is linearly separable~\cite{Shmilovici2010}. Figure \ref{fig:SVM} demonstrates the same input data in different dimensional feature spaces.

Once the decision boundary has been defined, new input data can be classified into the appropriate class depending on its position relative to the hyperplane. Advantages of using the SVM technique include the ability to work in higher dimensional feature spaces, which is beneficial for performing forecasts with many features. Solving the optimisation problem of the decision boundary also makes SVM models robust to outliers and overfitting. When training, the model attempts to maximise the distance between the hyperplane and the closest data points from each class, known as the support vectors, whilst also applying a cost for incorrect classifications. Since this cost decays with distance from the hyperplane, notable outliers do not have much effect on the decision boundary~\cite{LI2022491}.

Another prominent machine learning model is the Artificial Neural Network (ANN). Loosely inspired by the structure and function of the human brain, an ANN consists of layers of interconnected nodes (neurons), where each neuron takes input from one or more neurons in the previous layer, performs a simple computation (usually a weighted sum followed by a non-linear activation function), and passes the output to one or more neurons in the next layer. Using a training algorithm such as back-propagation, artificial neural networks can adjust the weights on the connections between neurons to approximate complex input-output mappings. Some common training algorithms are as follows:

\begin{itemize}
    \item CMA-ES: Evolutionary algorithm for continuous optimisation, adaptively models correlations between variables, effective in complex landscapes.
    \item Back-propagation: Neural network training algorithm, calculates gradients by propagating errors, adjusts weights and biases to minimise the loss function, widely used for training deep learning models.
    \item Adam~\cite{adam}: Optimisation algorithm for neural network training, combines adaptive learning rate and momentum, efficiently handles sparse gradients, accelerates convergence --- a popular choice for various deep learning tasks.
\end{itemize}

Through careful choice of the NN architecture and choosing the most appropriate training algorithm, an ANN can learn to accurately make predictions on new, unseen data~\cite{ABHISHEK2012311}.

\begin{figure}[H]
\centering
\includegraphics[width=0.9\textwidth]{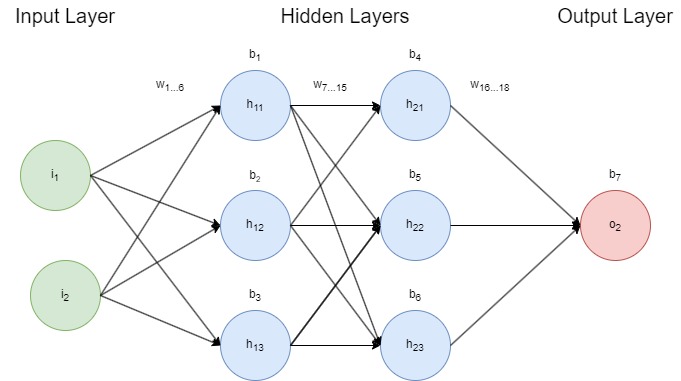}
\captionsetup{font=small}
\caption{An example of the structure of a simple neural network featuring 2 hidden layers}
\label{fig:neural}
\end{figure}

ANNs are powerful due to their ability to capture complex nonlinear relationships between input and output variables, making them well-suited for forecasting tasks where patterns in the data may be difficult to identify. Depending on the structure of the neural network, it is capable of capturing many obscure patterns and seasonalities without requiring expert knowledge of the subject matter. The challenge, however, lies in the fact that without experience in constructing complex ANNs, determining the optimal structure, training algorithms, and tuning methodology can be difficult. Depending on the purpose of the model, an ANN's lack of transparency and interpretability could pose an issue if information on how the model makes predictions is essential. % check this

The aforementioned ANN is typically referred to as a Feed-Forward Neural Network since the information flows in a single direction from input to output without and loops. A limitation of this architecture is the inability to capture temporal dependencies within the data. To assist with this, a more complex network design can be implemented which features a new \textit{recurrent connection} within each hidden node. This feedback loop allows for the network to better process sequential data by using the output from the previous step as an input to the current step. This type of architecture is known as a Recurrent Neural Network (RNN) and provides the benefit of allowing the model to better capture details of the current time step in context of previous time steps. This is particularly useful for handling sequential data such as in time-series forecasting.

\begin{figure}[H]
\centering
\includegraphics[width=0.6\textwidth]{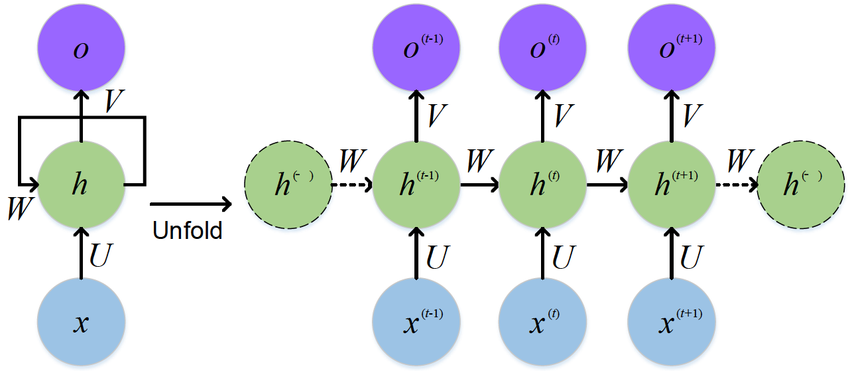}
\captionsetup{font=small}
\caption{Illustration of an RNN being unfolded across three time steps, where the recurrent connections are expanded to show the flow of information from input ($x$) to hidden state ($h$) and output ($o$) at each time-step, with weight matrices $U$, $W$ and $Z$ shared across all steps~\cite{rnn}.}
\label{fig:rnn}
\end{figure}

\begin{figure}[H]
\centering
\includegraphics[width=0.9\textwidth]{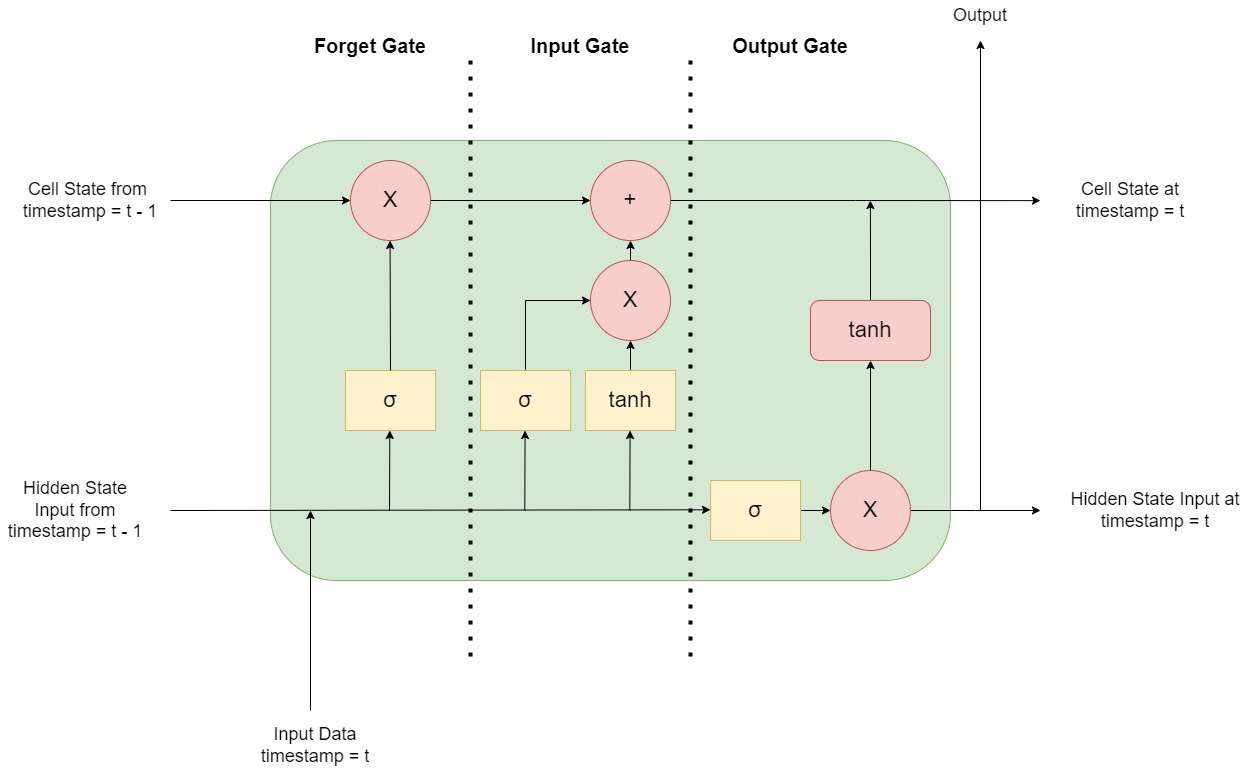}
\captionsetup{font=small}
\caption{An example of an LSTM cell featuring forget, input, and output gates}
\label{fig:lstm}
\end{figure}

Even though RNNs are more capable at capturing dependencies from sequential data sources, they struggle to fully capture dependencies which take a long time to appear due to the vanishing gradient problem~\cite{lstm}. The vanishing gradient problem refers to the issue in deep neural networks where the gradients diminish exponentially as they propagate backwards through layers, hindering the learning of long-term dependencies and resulting in ineffective weight updates. To help mitigate this issue, an architectural design called Long short-term memory (LSTM) can be used as shown in \ref{fig:lstm}.

The input gate, forget gate and output gate control the flow of information, allowing LSTMs to selectively retain and update relevant information over time. By learning to adjust these gate values during training, LSTMs can capture and retain important context from past time steps while disregarding irrelevant details. This mechanism enables LSTMs to make accurate predictions and model complex temporal patterns in tasks like language processing, or time series analysis.

Another popular technique which has applications within forecasting is Random Forest. Random Forest is a powerful machine learning algorithm that functions by combining the predictions from multiple decision trees to make a more accurate prediction. The ensemble of decision trees is generated from subsets of the training data using a process called \textit{bagging}. The outputs from all decision trees are aggregated to form a final prediction, often created from the average value of trees. Random forest techniques are popular due to their ability to handle large datasets of complex multidimensional data, capture non-linear relationships, and provide robustness against noise and outliers~\cite{AHMAD201777}. Even though these models can be powerful forecasting tools on their own if utilised properly, they are often combined in an attempt to harness their combined strengths and mitigate their weaknesses. These are known as hybrid models and typically consist of a combination of statistical and machine learning models~\cite{ann-rf, fuzzy-dt, forecasting2018}.

\begin{figure}[H]
\centering
\includegraphics[width=0.9\textwidth]{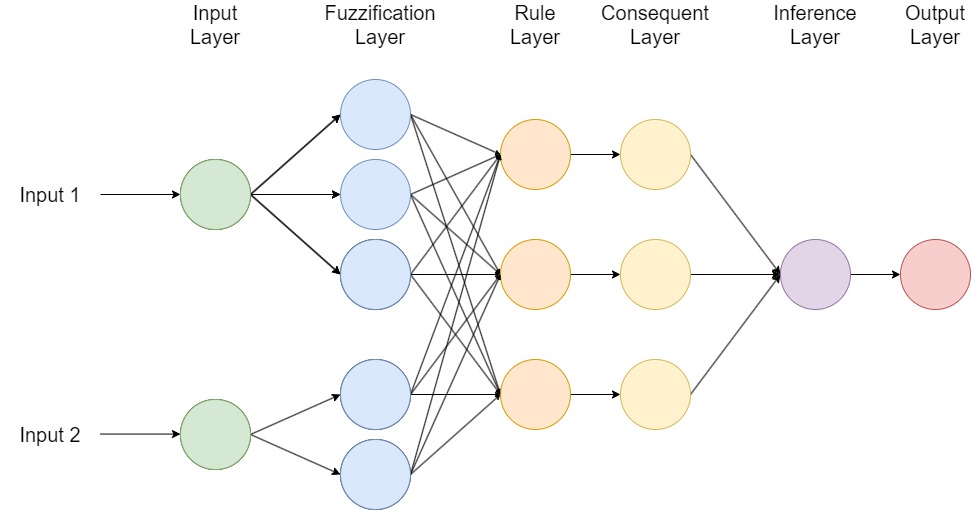}
\captionsetup{font=small}
\caption{A simple ANFIS architecture}
\label{fig:anfis}
\end{figure}

One of the most common hybrid models is the Adaptive Neuro-Fuzzy Inference System (ANFIS), which is a hybrid model that combines the learning capabilities of ANN and the interpretability of fuzzy logic. The ANFIS structure closely resembles the flow of a typical fuzzy model, with each stage of progression being represented by layers of neurons~\cite{ANFISforecast}. Figure \ref{fig:anfis} below demonstrates an example structure of an ANFIS system.

The ANFIS architecture typically consists of 5 layers. The first layer is the input layer, which receives input values and passes them to the next layer. The second layer is the membership function layer, or fuzzification layer, which calculates the membership value of the input variables to each fuzzy set. The third layer is the rule layer, which calculates the firing strength of each rule by taking the product of the membership values of the input variables. The fourth layer is the normalisation layer, which normalises the firing strengths. Finally, the fifth layer is the output layer, which aggregates the firing strengths of all the rules and generates the final output value~\cite{AComparativeStudyofNeuralNetworks2017}.

In the context of energy forecasting, ANFIS can be particularly useful due to its ability to model complex, non-linear relationships between variables such as weather conditions, historical energy consumption data, and other relevant factors. By continuously learning from new data, ANFIS can adapt to changes in patterns, improving the accuracy of its predictions over time.

% Picture Summarising all methods?

\section{Forecasting Techniques: Implementations}
\label{sec:Implementations}

When moving from theory to implementation, the effectiveness of each forecasting technique is heavily influenced by the exact parameters of the forecasting problem, especially the data and resources available. As demonstrated in this section, no single forecasting technique excels in all scenarios, especially when considering the desired forecasting horizon. The forecasting horizon denotes the time duration for which a forecast is made, representing a future period over which predictions are generated. Although names and definitions may vary, the commonly accepted forecasting horizons include:

\begin{itemize}
    \item \textbf{Very short-term}. From a second to 30 minutes.
    \item \textbf{Short-term}. From 30 minutes to 6 hours.
    \item \textbf{Medium-term}. From 6 hours and 1 day.
    \item \textbf{Long-term}. From 1 day to 1 week.
\end{itemize}

Knowledge of the required time horizon is extremely important when searching for an appropriate forecasting technique, and the choice is essentially an optimisation problem regarding the required responsiveness and accuracy of the model.

When exploring models for a particular forecasting problem, it is important to be able to determine whether the model has a suitable accuracy to meet the desired forecasting objectives and requirements. Many metrics can be employed to determine the accuracy of a forecasting model, a description of some of the most common metrics are as follows.

\begin{itemize}
    \item \textbf{Mean Absolute Error (MAE)~\cite{mining, inpatient, mae-wind, smape-astro}.} Measures the average absolute difference between the predicted values and the actual observed values. MAE is a simple and easy interpret metric and is robust to outliers in the data, however the inability to determine the direction and bias of the errors can cause an issue depending on the forecasting problem. 
    \begin{figure}[H]
    \[MAE = \frac{1}{n} \sum_{i=1}^{n} \left| y_i - \hat{y}_i \right|\]
    \end{figure}
    
    \item \textbf{Mean Squared Error (MSE)~\cite{mining, smape-astro}.} Calculates the average of the squared differences between the predicted values and the actual observed values. MSE is useful when more emphasis needs to be placed on larger errors, for example, if risk aversion is especially important. Due to the squaring nature of the metric, the error has different units from the original data which could potentially harm interpretability.
    \begin{figure}[H]
    \[MSE = \frac{1}{n} \sum_{i=1}^{n} (y_i - \hat{y}_i)^2\]
    \end{figure}
    \item \textbf{Root Mean Squared Error (RMSE)~\cite{mining, inpatient, smape-astro, mae-wind}.} Calculates the square root of the MSE. Useful for interpretability due to having the same units as the original data.
    \begin{figure}[H]
    \[RMSE = \sqrt{MSE}\]
    \end{figure}
    \item \textbf{Mean Absolute Percentage Error (MAPE)~\cite{mining, inpatient, smape-astro, mae-wind}} Calculates the average percentage difference between the predicted values and the actual observed values. Useful for when there is a need to compare accuracy across data sets with different scales since it provides a relative measure, however, the error metric can become undefined when the data contains zero values since the error will become undefined.
    \begin{figure}[H]
    \[MAPE = \frac{1}{n} \sum_{i=1}^{n} \left| \frac{y_i - \hat{y}_i}{y_i} \right| \times 100\]
    \end{figure}
    \item \textbf{Symmetric Mean Absolute Percentage Error (SMAPE)~\cite{smape-metric, smape-astro}} A variation of MAPE which divides by the sum of the forecast value and actual value, instead of just the actual value. This allows the metric to better deal with zeroes in the data compared to MAPE, although undefined values can still occur.
    \begin{figure}[H]
    \[SMAPE = \frac{1}{n} \sum_{i=1}^{n} \frac{\left| y_i - \hat{y}_i \right|}{(|y_i| + |\hat{y}_i|)/2} \times 100\]
    \end{figure}
\end{itemize}

In these equations, \(n\) represents the total number of data points, \(y_i\) represents the observed or actual values, and \(\hat{y}_i\) represents the corresponding forecast or predicted values. To summarise, the choice of metric is important and heavily dependent on the individual forecasting problem.

\subsection{Energy Generation Forecasting}

Mini-grids usually rely on renewable energy sources, such as wind and solar, and a backup diesel generator to provide power. Due to the weather dependency of the renewable assets, predicting the power available at a future time will require accurate information on future weather conditions, as well as the availability of the energy assets~\cite{irena2020}. 

\begin{figure}[H]
\centering
\includegraphics[width=0.8\textwidth]{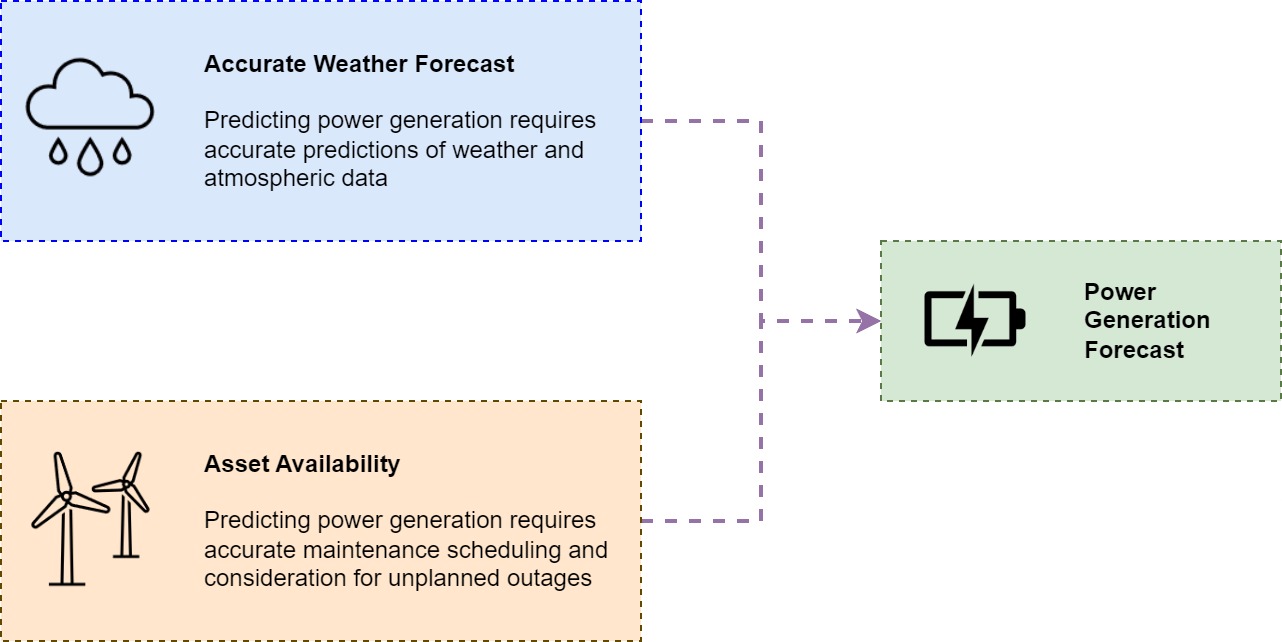}
\captionsetup{font=small}
\caption{Creating a power generation forecast}
\label{fig:powerforecast}
\end{figure}

As such, to have a viable mini-grid, it is important to accurately determine the energy supply in both the short-term and the medium/long-term. Short-term forecasts are highly beneficial for supporting near real-time supply and demand balancing, thus helping to maintain grid stability. Accurate long-term forecasts would provide the ability to provide more operational security in regards to ensuring sufficient energy reserves, along with allowing for more risk-free maintenance scheduling~\cite{soman2010review}.

Multiple statistical and intelligent forecasting models for short-term renewable energy generation (solar) and consumption were examined by Abu-Salih B et al.~\cite{abu-salih2022short}, and their empirical results were compared. The study collected real-time energy data from smart meters installed in residential premises in Western Australia covering energy import and export from the grid, rooftop photo-voltaic (PV) energy generation, household energy consumption, and outdoor temperature. Taking 11 months of data, splitting into 9 months training and 2 months testing, they pre-process the data through transformation into 1-hour intervals, as well as integrating separate temperature data from the Bureau of Meteorology for Perth, and adding a new day-of-week feature. For multivariate forecasting using supervised deep learning, the data was split into input features and an output label. The input features included temperature, day of the week, and energy consumption or generation at the previous time step, while the output label represented the current energy consumption or generation. Additional steps were taken to ensure that the data was stationary and appropriately scaled by calculating the min/max coefficients of all features. The team then performed extensive testing using a large number of linear/statistical models (ARIMA, Linear Regression), as well as a number of machine learning models (ANN, SVM, Random Forest), and a novel optimised LSTM model. The team observed that statistical techniques generally outperformed AI techniques in their study. For instance, Linear Regression achieved an approximate Mean Absolute Error (MAE) of 1.57 and Root Mean Square Error (RMSE) of 2.11 for energy generation (EG) and consumption (EC). In contrast, an Artificial Neural Network (ANN) model had an MAE of 8.687 for energy generation (EG) and 5.331 for energy consumption (EC), with RMSE values of 12.742 (EG) and 7.353 (EC). The optimised Long Short-Term Memory (LSTM) model developed by the team outperformed all other techniques, achieving an MAE of 0.329 (EG) and 0.241 (EC), and RMSE of 0.5654 (EG) and 0.3273 (EC).

Mason et al.~\cite{MASON2018705} examine forecasting energy demand, wind generation, and carbon dioxide emissions in Ireland using evolutionary neural networks. The study utilises the covariance matrix adaptation evolutionary strategy (CMA-ES) as an optimisation algorithm to train neural networks for short-term predictions and evaluate accuracy against other NN optimising strategies, as well as a small number of statistical models. The team chose to use recurrent neural networks (RNNs) over feed-forward networks due to their ability to retain memory of previous time steps, making them an appropriate choice for time-series forecasting problems. The RNN architecture used in the study contains six hidden neurons and two inputs representing the current and previous time steps within the time-series. The networks are fully connected, and their output corresponds to the prediction for the next time step. Seven algorithms, including CMA-ES, particle swarm optimisation (PSO), differential evolution (DE), back-propagation (BP), moving average (MA), random walk forecasting (RWF), and linear regression (LR), are evaluated for forecasting accuracy using four standard metrics: mean absolute error (MAE), mean squared error (MSE), root mean squared error (RMSE), and mean absolute percentage error (MAPE). Each network was trained on one month of data and tested against the following month of data. Each data set consists of time-series data in increments of 15 min, totalling 2878 data points. CMA-ES demonstrates superior performance in terms of convergence rate and accuracy compared to all other algorithms. It achieves the highest prediction accuracy across all three forecasting problems (wind generation, energy demand, carbon dioxide emissions). It was also noted that Linear Regression performs surprisingly well in predicting wind power generation but struggles with power demand due to its complex non-linearities and frequent steep gradients.

Chen, Gooi, \& Wang~\cite{CHEN2013195} introduce a solar radiation forecast technique that combines fuzzy logic and neural networks to achieve accurate predictions in various weather conditions. The proposed approach leverages future sky conditions and temperature information obtained from the National Environment Agency (NEA) to forecast solar radiation. The researchers emphasise the notable influence of factors such as sky condition, temperature, and time information on the received solar radiation by photovoltaic (PV) cells. To perform the forecasts, a feed-forward neural network (FNN) is employed, utilising historical solar radiation and weather data for training. The input layer comprises past solar radiation, temperature, and sky information, while the output layer predicts the solar radiation for the subsequent hour. During the training process, historical data is utilised for each hour of the daytime, classifying sky conditions into categories such as sunny, rainy, and cloudy for each hour. The temperature variable is employed to discern the differing solar radiation patterns under distinct sky conditions. The determination of parameters for each neural network is achieved through an algorithm, and approximately 21,600 data points are employed for training, validation, and testing purposes. The effectiveness of the proposed approach is validated through a case study encompassing four distinct scenarios (sunny, cloudy, rainy, and normal). The study demonstrates a significant reduction in the Mean Absolute Percentage Error (MAPE) compared to other existing solar radiation forecasting methods including an ARIMA model, as well as individual fuzzy and neural network models.

Li~\cite{Li2001} developed a neural network-based prediction model to estimate the power produced by individual turbines at the Fort Davis wind farm using data collected from each turbine. The authors use data collected from the Fort Davis wind farm, and develop a four-input neural network, using wind speed and direction from two sources, that outperforms a traditional single-parameter model. The study found that wind velocity has a significant impact on power output, while wind direction has less influence, particularly at higher power generation levels. A separate neural network was developed for each turbine to predict its performance, as this approach reduces the size and complexity of the neural network and scales better for large wind farms. The input-output mapping was modelled using a multi-layer perceptron, and the resulting neural network is shown to estimate wind power generation efficiently as a diagnostic tool, with the estimated power output closely matching the measured power output, in contrast to the traditional model's estimates.

In a 1998 study~\cite{ALEXIADIS199861}, the authors present a method for forecasting wind speed and electrical power generated by wind turbines, specifically for power systems with integrated wind parks. Artificial neural network models were used to forecast average values of wind speed and corresponding electrical power for 10-minute or 1-hour intervals using wind speeds and their derivatives as inputs. The team used data collected over seven years from six sites on islands of the South and Central Aegean Sea in Greece for testing. Auto-correlation patterns were discovered, revealing daily, monthly, and season-long periodicities. Recent wind speeds at the site and neighbouring wind speeds directed towards the site were found to be the best inputs for the model, leveraging spatial correlation. The team found better success using differenced wind patterns, and a simple ANN was used for forecasting, producing satisfactory results.

\subsection{Energy Demand Forecasting} % Maybe cut and combine

Accurately forecasting the energy demand of mini-grid consumers is essential for any mini-grid management solution. This accurate forecasting is crucial for determining whether the energy generated by the grid's renewable resources will be sufficient to meet demand at all times. This is especially important since the grid's backup diesel generators will most likely require preparation and startup time before it can effectively meet energy demands. Combined with energy generation forecasts, it will become easier to provide a more reliable grid and better forecast grid maintenance.

In a study by Bahrudin Hrnjica and Ali Danandeh Mehr~\cite{Hrnjica2020}, a specialised version of the recurrent neural network (RNN), namely long short-term memory (LSTM), is proposed as an efficient alternative for an energy prediction model in Nicosia, the capital of Cyprus. The proposed approach utilises a deep learning LSTM model combined with an auto-encoder for unsupervised learning. The model is fed with 15 input features, consisting of power demand values from previous time steps. The LSTM layer is followed by a 30\% dropout layer, and produces a single output. The LSTM model was trained and validated on data from January 2011 to November 2016, with the December 2016 values used as a test set. The study compared their LSTM deep learning model to a TSD (Time Series Decomposition) model, which utilised the Prophet~\cite{TaylorLetham2018} forecasting package for additive modelling and seasonal decomposition. The TSD model~\cite{tsd} incorporated various forecasting methods, such as ARIMA and exponential smoothing, to obtain the best possible model. The energy demand dataset from January 2011 to September 2015 was used for building the TSD model, while the period from October 2015 to December 2016 served as the validation and testing dataset. Results from their study showed that the LSTM deep learning model they developed outperformed the Prophet TSD model in various forecast horizons. Specifically, the LSTM model achieved lower root mean squared error (RMSE) values compared to the TSD model for 5-day forecasts (198 vs. 353), 2-week forecasts (320 vs. 386), and 1-month forecasts (281 vs. 720). This indicates the superiority of the LSTM approach in accurately predicting energy demand over multiple time intervals and the potential of LSTM-based approaches for improving energy prediction models in smart cities.

Tao Hong, Min Gui, Mesut E. Baran, and H. Lee Willis~\cite{hong} propose using multiple linear regression for modelling and forecasting short-term electric load. The paper explores the causal factors driving the fluctuations of electric load and their impacts on energy purchase and demand side management (DSM). The authors observe from the data that there are clear seasonalities within each day, week, and year, heavily influenced by temperature and human activities. The regression model proposed by the team was used to generate a 3-year hourly demand forecast and was shown to have significantly improved performance with a MAPE of 4.558\% over the existing mid-term load forecast used by an unspecific US utility company. %expand?

In a paper by Guan et al.~\cite{guan}, the authors explore Very Short-Term Load Forecasting (VSTLF) methods for predicting electric power system loads a few minutes into the future using Wavelet Neural Networks trained by Hybrid Kalman Filters (WNNHKF). The team incorporates wavelet analysis and hybrid Kalman filters into the model which forecast loads one hour into the future in 5-minute steps using a moving window approach. The performance of different models was evaluated in terms of Mean Absolute Percentage Error (MAPE) for 5-minute and 60-minute load forecasts:

\begin{itemize}
    \item Persistence: 0.38 MAPE (5 minutes), 4.29 MAPE (60 minutes)
    \item Linear AR: 0.16 MAPE (5 minutes), 1.73 MAPE (60 minutes)
    \item WNN: 0.08 MAPE (5 minutes), 0.49 MAPE (60 minutes)
    \item WNNHKF: 0.12 MAPE (5 minutes), 0.47 MAPE (60 minutes)
\end{itemize}

The results demonstrate that the WNNHKF model developed by the team outperforms other methods, achieving lower MAPE values for both short-term and one-hour load forecasts.

Ahmad et al.~\cite{AHMAD201777} developed several machine learning models for predicting the energy consumption of a large hotel in Madrid. They compared the performance of a feed-forward back-propagation artificial neural network (ANN) with a random forest (RF) model. The study utilised a dataset of 5-minute historical values of Heating, Ventilation and Air Conditioning~(HVAC) electricity consumption, along with weather conditions and hotel occupancy data. The results indicated that the ANN model outperformed the RF model on various performance metrics with a MAPE of 4.97 vs 6.03 for the RF model. However, the team noted that the RF model demonstrated the ability to handle missing values effectively due to its inherent ensemble-based nature. Overall, they conclude that both models showed strong non-linear mapping generalisation ability and could be effective in predicting energy consumption in the problem area.

Allee et al.~\cite{ALLEE202156} conducted a study on estimating electricity demand for un-electrified customers in the context of the design of mini-grid systems. The team notes that accurately estimating electricity demand is crucial for site selection, revenue projection, and sizing system components in order to minimise capital costs. The researchers tested a data-driven approach using survey and smart meter data from 1378 Tanzanian mini-grid customers. Their findings revealed that machine learning models incorporating customer survey data consistently outperformed baseline models. The best-performing model, LASSO regression, achieved a median absolute error of 66\% and 37\% for individual connections and mini-grid sites, respectively. The study highlights the importance of thorough inventories of customers' currently-owned appliances rather than relying on detailed demographic information or self-reported habits and plans. They conclude by stating that by combining shortened questionnaires with smart meter data, accurate estimates of initial customer electricity demand can be significantly improved compared to standard field practices.

A study conducted by Cevic and Cunkas~\cite{Cevik2015} involved short-term load forecasting using fuzzy logic and adaptive neuro-fuzzy inference system (ANFIS). The team analysed several years of historical data and noted the unique load characteristics of different parts of the week. The inputs for their ANFIS forecasting model included historical load, temperature difference, and season. Unlike previous studies, they performed hourly load forecasts for an entire year and demonstrated that fuzzy logic yielded good results for large datasets over an extended period. The study utilised electricity consumption data from the Turkish Electricity Transmission Company (TEIAS), for the years 2009 to 2012. The authors emphasised the influence of time factors on load curves, which vary with season, week, day, and hour. Meteorological variables, particularly temperature, were identified as crucial factors affecting short-term load forecasting. The study proposed fuzzy logic and ANFIS models for load forecasting and compared their results with actual load data. The inputs considered for the forecasting system included historical load, temperature difference, and season. Season was included because changes in temperature can have opposite effects on load depending on the time of year. The results indicated that both ANFIS and fuzzy logic methods produced accurate forecasts, with the ANFIS model exhibiting better forecasting accuracy, as evidenced by lower mean absolute percentage errors (MAPEs) of 1.85\% compared to 2.1\% for fuzzy logic, based on data from the year 2012.

\subsection{Energy Management Forecasting}

The forecasting of mini-grid management and operations plays a crucial role in ensuring the reliable and efficient functioning of mini-grid systems. Accurate operation forecasts would enable mini-grid operators to optimise the deployment and scheduling of all energy resources, better plan maintenance activities, ensure grid stability and make informed decisions regarding energy procurement and storage capacity. Reliable forecasting would minimise operational costs, lower carbon emissions by reducing reliance on diesel backup generators, and ultimately enhance the overall reliability and sustainability of the mini-grid system. 

Chaouachi, Kamel, Andoulsi, and Nagasaka~\cite{6157610} demonstrate research into solving multi-objective intelligent energy management problems within micro-grids, especially addressing decision problems with conflicting objectives. The optimal solution aims to minimise the operation cost and gaseous emissions while maintaining load balance within the power generation limits of each controllable distributed generator (DG). The cost objective function considers power generation costs, energy trade between the micro-grid and the main grid, and maintenance costs proportional to DG power generation and storage. The emission objective function incorporates individual DG emission rates during autonomous generation and the average emission rate of the main grid in case of a generation shortfall, as well as the fact that selling power to the grid reduces overall emissions. The team note that battery scheduling is an integral part of achieving the optimal energy management objectives. An expert system based on fuzzy logic is developed for micro-grid battery scheduling, complemented by multi-objective linear programming optimisation for cost and emission considerations. The fuzzy logic system relies on seven measured and forecast parameters designed using expert knowledge. These parameters include time, battery state of charge, electricity price, load demand, renewable energy generation, and weather forecasting. The team then defined a set of fuzzy rules to determine optimal battery charge or discharge rates based on the state of the aforementioned parameters. As an example, charging prioritises low load demand periods and high renewable energy availability, while discharging aims to meet load balance without exceeding depth-of-discharge limits or requiring expensive energy purchases. The proposed methodology demonstrates significant cost savings during peak load demand periods and non-availability of solar generation. Simulation results show an approximate 4.5\% cost savings and 3\% lower emissions compared to conventional approaches, highlighting the effectiveness of the proposed multi-objective intelligent energy management system.

\begin{table}[t!]
  \centering
  \label{tab:modelsummary}
  \begin{tabular}{|p{2.5cm}|p{1cm}|p{2.5cm}|p{5.5cm}|}
    \hline
    Forecast & Source & Model Type & Features \\
    \hline
    \multirow{3}{=}{Energy Generation Forecasting} & \cite{MASON2018705} & RNN using CMA-ES & Current and previous time-steps \\
    \cline{2-4}
    & \cite{Li2001} & ANN & Wind speed and direction \\
    \cline{2-4}
    & \cite{ALEXIADIS199861} & ANN & Wind speed and direction \\
    \cline{2-4}
    & \cite{6157610} & RNN & Max wind speed, past 3 wind speed, wind direction, pressure, humidity\\
    \hline
    \multirow{3}{=}{Energy Supply Forecasting} & \cite{CHEN2013195} & ANFIS & Past solar radiation, temperature, sky information \\
    \cline{2-4}
    & \cite{6157610} & RNN & Max temp, min temp, temp, pressure, precipitation, solar irradiance\\
    \cline{2-4}
    & \cite{abu-salih2022short} & Optimised LSTM & Temperature, day of the week, energy generation at previous time-step\\

    \cline{2-4}
    & \cite{Hrnjica2020} & LSTM + Auto-encoder & 15 previous energy demand time-steps \\
    \cline{2-4}
    & \cite{guan} & WNNHKF & 12 previous energy demand time-steps\\ 
    \cline{2-4}
    & \cite{AHMAD201777} & FFNN & current energy demand, weather conditions, hotel occupancy \\
    \cline{2-4}
    & \cite{ALLEE202156} & LASSO regression & Smart meter data, customer asset estimation \\
    \cline{2-4}
    & \cite{Cevik2015} & ANFIS & Historical load data, temperature, season\\
    \hline
    \multirow{2}{=}{Energy Management Forecasting} & \cite{6157610} & Fuzzy Logic &  Time, battery state of charge, electricity price, load demand, renewable energy generation, and weather forecasting\\
    \hline
  \end{tabular}
    \caption{A summary of existing research and models}
\end{table}

\section{Overview and Recommendations}

% To attempt to provide a more thorough evaluation of the aforementioned forecasting models, with further exploration of model evaluation techniques, this section will attempt an introductory implementation of various forecasting models on a test data set.

In this section, we implement two time-series forecasting techniques identified as appropriate during our earlier analysis of the literature. We also discuss the utility of open-source energy demand and supply forecasting datasets for model exploration and transfer learning.

\subsection{Public Datasets for Testing}
Since it will take time to install and generate real data from the project mini-grid, any model development will require large amounts of relevant real-world data or at least accurate simulated data.

\subsubsection{Electrical consumption, generation, pricing, and weather data from Spain}

This dataset, available at \url{https://www.kaggle.com/datasets/nicholasjhana/energy-consumption-generation-prices-and-weather}, contains 4 years of electrical consumption, generation, pricing, and weather data for Spain with a 1-hour frequency. The data is contained within two files \textit{energy\_dataset.csv} and \textit{weather\_features.csv}, each of which is described in more detail below.

\textbf{energy\_dataset.csv}
Contains 29 features including the amount of energy in MW from a variety of energy sources such as coal, wind, solar, hydro, and gas, along with the total energy load at that time and the price of energy at that time.

\textbf{weather\_features.csv}
Contains 17 features including the min/max/current temperature, pressure, humidity, wind speed and direction, for the 5 largest cities in Spain.

The dataset's available features correspond to the inputs commonly used in energy generation and consumption models studied during the research. However, it is important to consider potential challenges associated with the data. One such challenge stems from the dataset covering a broad geographical area, rather than a localised mini-grid setting. Additionally, variations in climate trends between Spain and the target country, as well as distinct cultural practices and holidays specific to each region, may pose additional considerations.

The dataset is also potentially limited by the lack of information on Carbon emissions from the various energy sources, as well as information regarding energy trading prices. Since energy forecasting involves a multi-objective optimisation problem aimed at balancing cost, carbon footprint, and hardware operations and maintenance, additional external information needs to be incorporated into this dataset to achieve these objectives.

\subsubsection{Appliances Energy Prediction}
The ``Appliance Energy Prediction'' dataset available on the UCI repository~\cite{misc_appliances_energy_prediction_374} can be used to fit forecasting models for energy usage in household appliances. The dataset focuses on the energy consumption and prediction for a low-energy house located in Stambruges, Belgium, approximately 24 km from the City of Mons. The house was completed in December 2015 and features new mechanical systems. Designed according to passive house certification standards, the house aims for an annual heating and cooling load of no more than 15 kWh/m² per year, as determined by the Passive House Planning Package (PHPP) software.

Key features of the house include a wood chimney for heating, with monthly logs of the type and amount of wood used, and a measured air leakage rate of 0.6 air changes per hour at 50 Pa. The house has highly insulated exterior walls, roof, and ground (U < 0.1 W/m² K), and triple-glazed windows (Ug = 0.5 W/m² K, Uf < 0.9 W/m² K). Ventilation is provided by a heat recovery unit with 90-95\% efficiency. The total floor area is 280 m², with a heated area of 220 m². The house's façade is oriented +10° (Southwest) from due South. Typically, the house is occupied by four individuals—two adults and two teenagers—with one adult working regularly from a home office.

The dataset includes 29 features related to household energy consumption. Key variables include timestamps, energy use for appliances and lights, temperature and humidity measurements in different rooms, and weather conditions such as outdoor temperature, atmospheric pressure, humidity, wind speed, visibility, and dew point. Additionally, two random variables are included for regression model testing. 

For energy demand modelling, this dataset could be a valuable asset for benchmarking the efficacy of various forecasting techniques. Additionally, it could prove a useful source to extract fundamental energy usage patterns which can later be used to improve target models through transfer learning.

\subsection{Building Energy Efficiency}
The Energy Efficiency Dataset~\cite{misc_electricityloaddiagrams20112014_321} aims to predict the heating and cooling load requirements for buildings. It includes features such as relative compactness, surface area, wall area, roof area, overall height, and orientation. Furthermore, it accounts for the glazing area and its distribution. These features reflect architectural and environmental characteristics influencing a building's energy efficiency. The target variables in the dataset are the heating and cooling loads, which are critical for understanding and improving energy consumption patterns in buildings.

This dataset is highly relevant for optimising mini grids by enabling the development of predictive models for energy distribution and storage. Accurate predictions of heating and cooling loads help in planning and balancing the energy demand within mini grids. This ensures efficient utilisation of resources, supports the integration of renewable energy sources, and enhances the sustainability and reliability of energy supply in small-scale power systems.

\subsection{Smart Grid Smart Meter}
The Smart Grid Smart Meter Data dataset~\cite{misc_bias_correction_of_numerical_prediction_model_temperature_forecast_514} is designed for the purpose of bias correction of next-day maximum and minimum air temperature forecasts from the Local Data Assimilation and Prediction System (LDAPS) model~\cite{ldaps} operated by the Korea Meteorological Administration. The dataset contains fourteen numerical weather prediction (NWP) meteorological forecast variables, two in-situ observations, and five geographical auxiliary variables. It covers summer data from 2013 to 2017 over Seoul, South Korea. The input data includes next-day forecasts from the LDAPS model, present-day in-situ maximum and minimum temperatures, and various geographic variables. The outputs are the next-day maximum and minimum air temperatures. Hindcast validation --- statistical calculations determining probable past conditions --- was conducted for the period from 2015 to 2017.

The dataset comprises weather station data, present-day temperature records, and various meteorological forecasts, such as relative humidity, temperature lapse rates, wind speed, latent heat flux, cloud cover, and precipitation for different periods. It also incorporates geographic features like latitude, longitude, elevation, slope, and solar radiation.

By leveraging this dataset, mini grid operators can refine their energy management strategies. Accurate temperature predictions help in forecasting energy demand, crucial for balancing supply and ensuring efficient energy storage. This optimisation is vital for integrating renewable energy sources and enhancing the sustainability and reliability of mini grids. The dataset's comprehensive coverage of weather and geographic variables allows for detailed analysis and improvement of energy usage predictions, leading to better resource allocation and a more resilient energy system.

\subsection{Examination of Additional Public Tools}
As well as the existing research covered within Section \ref{sec:Implementations}, an examination was also performed on a small number of existing tools and software which could help facilitate forecasting. 

\subsubsection{Prophet}
An article by Taylor and Letham~\cite{TaylorLetham2018} describes what they call ``forecasting at scale,'' where they design a highly configurable decomposable time-series model. This model features interpretable parameters such as trend, seasonality, and holidays, which can be easily understood and adjusted using expert knowledge. Made available through the Facebook Open Source project, the Prophet forecasting model is easily accessible through a Python installation and provides the following features:

\begin{itemize}
    \item Prophet models the trend component using a piece-wise linear function allowing it to accurately model changes in trends over time.
    \item Complex seasonal components such as daily, weekly, or yearly patterns can be captured using Fourier series analysis, and custom seasonalities can be added if known.
    \item Prophet can handle holiday components that are affecting the time series, such as national holidays, using a built-in library of holidays for many countries and cultures.
    \item Provides interactive forecasts very quickly.
\end{itemize}

\begin{figure}[H]
\centering
\includegraphics[width=0.8\textwidth]{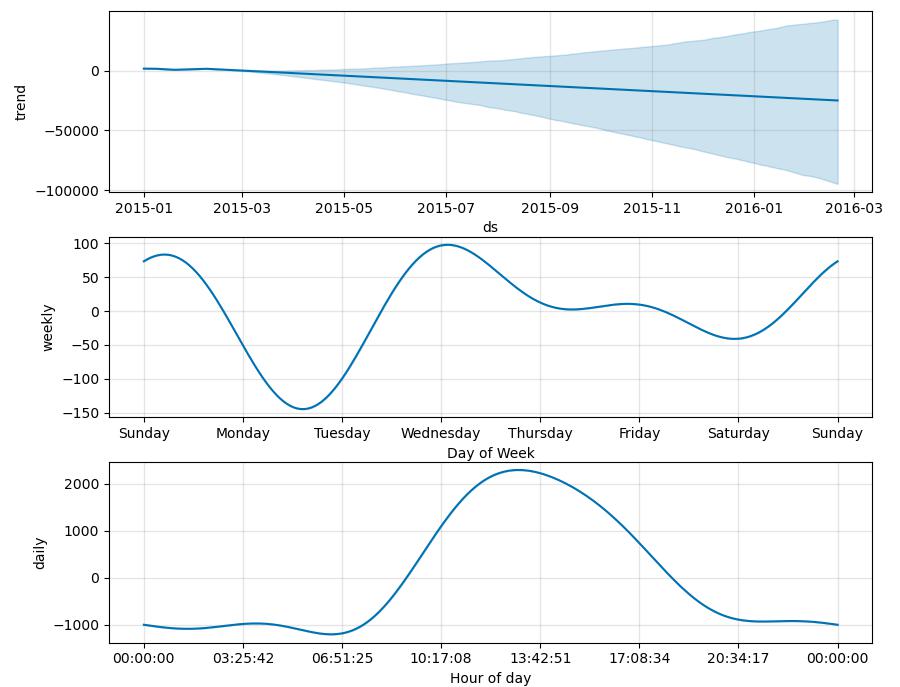}
\captionsetup{font=small}
\caption{Initial examination of the solar energy test-data using Prophet time-series decomposition}
\label{fig:prophettrends}
\end{figure}

Figure \ref{fig:prophettrends} below demonstrates the capability of the Prophet forecasting tool in automatically determining trends within the data. Prophet was assigned the task of analyzing a one-year dataset of solar energy generation and identifying the variations in energy generation throughout the day and week.

\subsubsection{NeuralProphet}

NeuralProphet is an open-source time series forecasting library developed by Facebook's Core Data Science team. It is built on top of the popular PyTorch library and provides a user-friendly interface for creating and training neural network models specifically designed for time series forecasting tasks. NeuralProphet extends the traditional Prophet forecasting model by incorporating a neural network architecture into its framework, allowing it to better capture complex nonlinear relationships and patterns in time series data. Key features of NeuralProphet include:

\begin{itemize}
    \item NeuralProphet allows users to define various components of a time series model, such as trend, seasonality, and events, providing flexibility in modelling different types of time series data.
    \item NeuralProphet has built-in mechanisms to handle missing values in time series data, reducing the need for manual preprocessing steps.
    \item NeuralProphet incorporates a neural network model based on Long Short-Term Memory (LSTM) units, enabling it to capture long-term dependencies and nonlinear patterns in time series data.
\end{itemize}

For short forecasts, NeuralProphet claims to improve forecast accuracy compared to Prophet by 55 to 92 percent~\cite{DBLP:journals/corr/abs-2111-15397}.

\subsubsection{N-BEATS}

N-BEATS (Neural basis expansion analysis for interpretable time series forecasting) is a state-of-the-art deep learning model for time series forecasting. It was introduced by Oreshkin et al. in 2020. N-BEATS is designed to capture complex temporal patterns in time series data and provide accurate and interpretable forecasts~\cite{DBLP:journals/corr/abs-1905-10437}.

N-BEATS works by decomposing a time series into a set of basic patterns called "basis functions" which represent different types of patterns that can occur in the data, such as trends, seasonality, or fluctuations. N-BEATS uses a neural network architecture, composed of fully connected (FC) layers, to learn the coefficients and weights associated with each basis function. These coefficients determine the importance or contribution of each basis function at different points in time. It starts with a large set of basis functions and progressively selects the most influential ones, removing the less important ones during training. N-BEATS offers interpretability by providing insights into the underlying patterns. The learned basis functions can be visualised to understand which patterns are driving the forecasts, helping users gain insights into the data.

\begin{figure}[H]
\centering
\includegraphics[width=1\textwidth]{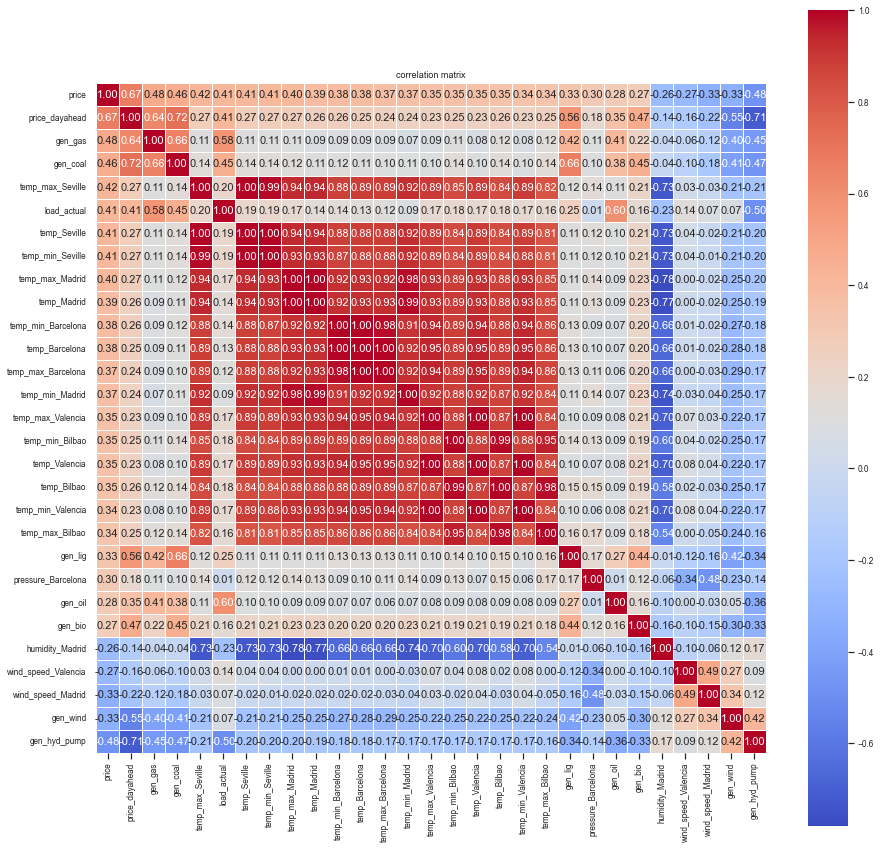}
\captionsetup{font=small}
\caption{Correlations between the features of the test dataset}
\label{fig:nbeatscor}
\end{figure}

As an example, the N-BEATS tool allows for quick analysis of the features of the dataset to determine the correlation between the features. Figure \ref{fig:nbeatscor} below is the result of a brief examination into the \textit{weather\_features.csv} dataset to see how the features are connected. This information will be useful in determining which features are the most important to explore within future forecasting models.

\subsection{Future Challenges and Recommendations}

When analysing energy generation, it is important to recognise that wind and solar energy each exhibit their own trends and characteristics. The extent to which these trends are shared will need to be established, as well as how each is influenced by factors such as temperature and general weather conditions. Simple regression and decomposition techniques such as ARIMA or Prophet are powerful tools that can be used to establish trends in historical data for each energy source individually and as such will be a good starting point. However, the correlation of the output from these energy sources with attributes such as temperature cannot be ignored. Higher temperatures generally enhance solar energy production due to increased sunlight intensity and duration, whilst wind energy production is influenced heavily by temperature gradients. Advanced techniques such as RNNs or LSTMs could be considered to tackle these dependencies, improve accuracy, and capture long-term patterns within the data. Issues which need to be explored within this model include determining which features are the most important when making forecasts. There are a great many atmospheric variables which can be considered, and determining the most influential will require a greater exploration of the data.

When exploring the forecasting of energy consumption, two suitable models include Prophet and the deep learning model Neural Prophet. These models offer several advantages that make them ideal for further exploration in energy consumption forecasting. Both tools allow for easier examination of trends, seasonalities, and the impact of holidays, all of which heavily affect user behaviour and significantly impact energy consumption. A benefit to this approach is that it allows for the exploration of non-intelligent time-series decomposition techniques which from research have been shown to be surprisingly effective in certain problem areas, whilst allowing for a relatively easy transition to the deep-learning model extension. An appropriate third alternative for exploration is an optimised LSTM model using a large number of previous time-step data since the LSTM architecture will allow for the capture of longer-term dependencies within time-series data. A challenge to overcome when forecasting energy demand will be finding an effective way to account for the nuanced cultural behaviours of the mini-grid community. Maximising the reliability of the grid requires knowledge of trends in behaviour, which might not become apparent until real data can be analysed. Therefore, it is crucial to explore techniques that enable the selected models to effectively handle unforeseen events and scenarios.

One of the best techniques to explore for handling the optimisation of multiple objectives, as required for operation and management forecasting, is fuzzy logic inference. Fuzzy logic is particularly suitable for this problem due to the numerous features which will need to be to considered, as it can effectively handle the uncertainty associated with each of these inputs using a set of established fuzzy rules. To optimise hardware longevity, low cost, low carbon emissions, and low risk, features such as time, battery state, electricity buy and share price, load demand, renewable energy generation, and weather forecasting will need to be considered. A simple fuzzy logic model could account for all of these features whilst factoring in their associated uncertainties in order to forecast optimal times to charge and discharge the battery, perform maintenance, connect to the main grid to share energy, and prepare for using the diesel generator. Exploring the hybrid ANFIS model would be advantageous for leveraging the interpretability and linguistic reasoning inherent in fuzzy logic, alongside the learning and adaptation capabilities offered by neural networks. A core issue with this model will be determining how to appropriately weigh the optimisation objectives based on a community's needs.

Another significant issue is the substantial delay in obtaining real data for both training and evaluation purposes. This delay presents a notable challenge as it raises uncertainties about the effectiveness of models trained solely on test data when applied to real-world scenarios. To address this concern, additional exploration of knowledge transfer techniques is essential. A promising approach to consider is the integration of adapter layers into the model architecture. Adapter layers can improve the models' adaptability to different data domains and enhance their performance on real data. This and other knowledge transfer techniques must be explored for energy forecasting projects.

Exploring the problem of selecting an appropriate forecast horizon is essential as it requires striking a balance between capturing long-term trends and mitigating excessive uncertainty. It will be important to thoroughly examine the chosen models across different forecasting horizons and consider data constraints to ensure their ability to accurately forecast within useful time frames whilst minimising associated risks.

%%%%%% SC: Please do not remove this section or the file %%%%%%

% \newpage
% \input{av-update}
% \newpage

%%%%%% SC: Please do not remove this section or the file %%%%%%

\section{Conclusion}
In conclusion, this survey has highlighted the significant potential of AI-driven solutions to enhance the operation and sustainability of mini-grids in rural communities. The integration of various forecasting techniques, including statistical models, machine learning algorithms, and hybrid approaches, has demonstrated the capability to address the inherent unpredictability of renewable energy sources. Accurate forecasting of energy supply and demand is crucial for optimising mini-grid operations and ensuring reliable and cost-effective energy distribution.

The exploration of tools such as Prophet, NeuralProphet, and N-BEATS has shown promising results in improving forecast accuracy and interpretability. These tools provide valuable insights into trends, seasonality, and the impact of external factors such as weather and holidays on energy consumption and generation. The examination of public datasets has further emphasized the importance of real-world data in developing and validating forecasting models. However, challenges remain in adapting these models to specific local conditions, accounting for cultural behaviours, and handling the delay in obtaining real data for training and evaluation purposes.

Future research should focus on enhancing model adaptability through techniques like knowledge transfer and the integration of adapter layers. Additionally, the optimisation of forecasting horizons and the incorporation of fuzzy logic inference for multi-objective optimisation are critical areas for further exploration. By addressing these challenges, AI-driven mini-grid solutions can significantly contribute to achieving sustainable and reliable energy access for rural communities, thereby promoting economic development and improving the quality of life.

Overall, this survey underscores the importance of continuous innovation and adaptation in forecasting techniques to meet the evolving needs of mini-grid systems. The collaborative efforts of researchers and industry partners are essential in driving forward the development of robust and efficient mini-grid solutions.

\section*{Acknowledgements}
The authors wish to express their heartfelt appreciation to Andrew McLeman for his substantial contributions to this work. Andrew played a crucial role in the literature review, significantly influencing the scope and depth of this survey. Regrettably, he is no longer part of our team, and thus we are unable to list him as a co-author. This project received funding from the UKRI Energy Catalyst fund (Project Reference: 10040475), and we sincerely acknowledge their generous support.

\bibliographystyle{plain}
\bibliography{main}

%\appendix
%\section{Appendix}
%\subsection{Forecasting Model Evaluation}
% potentially add the further examinations here in the appendix, but removed for now

\end{document}